\documentclass[sigplan,10pt]{acmart}

\acmConference[EMC\textsuperscript{2}]{THE 1ST WORKSHOP ON ENERGY EFFICIENT MACHINE LEARNING AND COGNITIVE COMPUTING FOR EMBEDDED APPLICATIONS (EMC²)}{2018}{VA, USA}
\acmYear{2018}
\acmISBN{} 
\acmDOI{} 
\startPage{1}

\setcopyright{none}

\usepackage{booktabs}   
\usepackage{subcaption} 
\usepackage{color} 
\usepackage{graphicx}
\usepackage{nicefrac}
\usepackage{array}
\usepackage{libertine}
\usepackage{xcolor}
\usepackage{soul}
\usepackage{caption}
\begin{document}

\title{A Quantization-Friendly Separable Convolution for MobileNets}

\author{Tao Sheng}
\email{tsheng@qti.qualcomm.com}          
\affiliation{
	\institution{Qualcomm Canada Inc.}           
}

\author{Chen Feng}
\email{chenf@qti.qualcomm.com}          
\affiliation{
	\institution{Qualcomm Canada Inc.}           
}

\author{Shaojie Zhuo}
\email{shaojiez@qti.qualcomm.com}          
\affiliation{
	\institution{Qualcomm Canada Inc.}           
}

\author{Xiaopeng Zhang}
\email{parker.zhang@gmail.com}          
\affiliation{
	\institution{\emph{Q}ualcomm Canada Inc.}           
}

\author{Liang Shen}
\email{liang.shen@qti.qualcomm.com}          
\affiliation{
	\institution{Qualcomm Canada Inc.}           
}

\author{Mickey Aleksic}
\email{maleksic@qti.qualcomm.com}          
\affiliation{
	\institution{Qualcomm Technologies, Inc.}           
}
\begin{abstract}
As deep learning (DL) is being rapidly pushed to edge computing, researchers invented various ways to make inference computation more efficient on mobile/IoT devices, such as network pruning, parameter compression, and etc. Quantization, as one of the key approaches, can effectively offload GPU, and make it possible to deploy DL on fixed-point pipeline. Unfortunately, not all existing networks design are friendly to quantization. For example, the popular lightweight MobileNetV1 \cite{mobilenetv1}, while it successfully reduces parameter size and computation latency with separable convolution, our experiment shows its quantized models have large accuracy gap against its float point models. To resolve this, we analyzed the root cause of quantization loss and proposed a quantization-friendly separable convolution architecture. By evaluating the image classification task on ImageNet2012 dataset, our modified MobileNetV1 model can archive 8-bit inference top-1 accuracy in $68.03\%$, almost closed the gap to the float pipeline.
\end{abstract}



\maketitle

\section{Introduction}
\begin{figure}[t]
	\includegraphics[width=8.5cm]{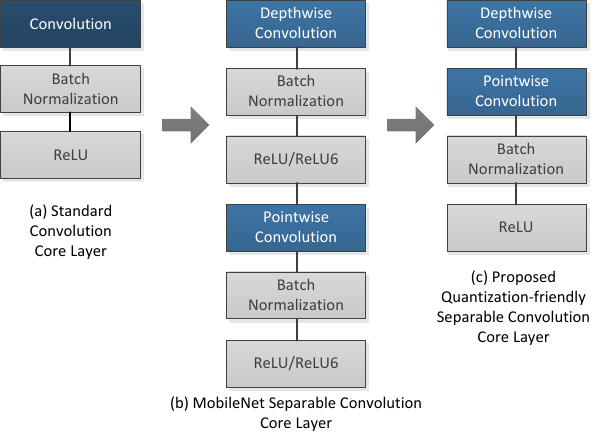}	
	\caption{Our proposed quantization-friendly separable convolution core layer design vs. separable convolution in MobileNets and standard convolution}
	\label{fig:sepconv}
	\centering
\end{figure}

Quantization is crucial for DL inference on mobile/IoT platforms, which have very limited budget for power and memory consumption. Such platforms often rely on fixed-point computational hardware blocks, such as Digital Signal Processor (DSP), to achieve higher power efficiency over float point processor, such as GPU. On existing DL models, such as VGGNet \cite{VGGNet}, GoogleNet \cite{GoogleNet}, ResNet \cite{ResNet} and etc., although quantization may  not impact inference accuracy for their over-parameterized design, it would be difficult to deploy those models on mobile platforms due to large computation latency. Many lightweight networks, however, can trade off accuracy with efficiency by replacing conventional convolution with depthwise separable convolution, as shown in the Figure \ref{fig:sepconv}(a)(b). For example, the MobileNets proposed by Google, drastically shrink parameter size and memory footprint, thus are getting increasingly popular in mobile platforms. The downside is that the separable convolution core layer in MobileNetV1 causes large quantization loss, and thus resulting in significant feature representation degradation in the 8-bit inference pipeline.

To demonstrate the quantization issue, we selected TensorFlow implementation of MobileNetV1 \cite{mobilenetmodel} and InceptionV3 \cite{gv3}, and compared their accuracy on float pipeline against $8$-bit quantized pipeline. The results are summarized in Table\ref{tab:mobilenettf8}. The top-$1$ accuracy of InceptionV3 drops slightly after applying the $8$-bit quantization, while the accuracy loss is significant for MobileNetV1.
\begin{center}
	\captionof{table}{Top-1 accuracy on ImageNet2012 validation dataset} \label{tab:mobilenettf8}
	\renewcommand{\arraystretch}{1.1}
	\begin{tabular}{|m{2cm}|m{1.2cm}|m{1.2cm}|m{2.5cm}|}
		\hline
		Networks	& Float Pipeline 	& 8-bit Pipeline 	& Comments \\ \hline \hline
		InceptionV3 & 78.00\% 			& 76.92\% 			& \small{Only standard convolution} \\ \hline
		MobileNetV1 & 70.50\% 			& 1.80\%  			& \small{Mainly separable convolution} \\ \hline
	\end{tabular}
\end{center}

There are a few ways that can potentially address the issue. The most straight forward approach is quantization with more bits. For example, increasing from $8$-bit to $16$-bit could boost the accuracy \cite{dl-compression-survey}, but this is largely limited by the capability of target platforms. Alternatively, we could re-train the network to generate a dedicated quantized model for fixed-point inference. Google proposed a quantized training framework \cite{quanttraining} co-designed with the quantized inference to minimize the loss of accuracy from quantization on inference models. The framework simulates quantization effects in the forward pass of training, whereas back-propagation still enforces float pipeline. This re-training framework can reduce the quantization loss dedicatedly for fixed-point pipeline at the cost of extra training, also the system needs to maintain multiple models for different platforms.

In this paper, we focus on a new architecture design for the separable convolution layer to build lightweight quantization-friendly networks. The proposed new architecture requires only single training in the float pipeline,  and the trained model can then be deployed to different platforms with float or fixed-point inference pipelines with minimum accuracy loss. To achieve this, we look deep into the root causes of accuracy degradation of MobileNetV1 in the 8-bit inference pipeline. And based on the findings, we proposed a re-architeched quantization-friendly MobileNetV1 that maintains a competitive accuracy with float pipeline, but a much higher inference accuracy with a quantized 8-bit pipeline. Our main contributions are:
	
\begin{enumerate}
	\item We identified batch normalization and ReLU6 are the major root causes of quantization loss for MobileNetV1.
	\item We proposed a quantization-friendly separable convolution, and empirically proved its effectiveness based on MobileNetV1 in both the float pipeline and the fixed-point pipeline.
\end{enumerate}

\section{Quantization Scheme and Loss Analysis}
\begin{figure}[t!]
\includegraphics[width=9cm]{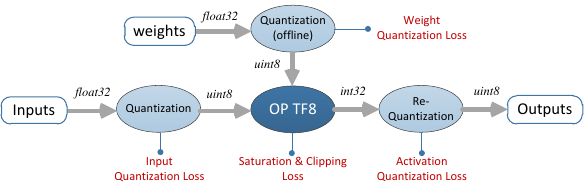}	
\caption{A fixed-point quantized pipeline}
\label{fig:quant-pipeline}
\centering
\end{figure}

In this section, we will explore the TensorFlow (TF) \cite{tf} $8$-bit quantized MobileNetV1 model, and find the root cause of the accuracy loss in the fixed-point pipeline. Figure \ref{fig:quant-pipeline} shows a typical $8$-bit quantized pipeline. A TF $8$-bit quantized model is directly generated from a pre-trained float model, where all weights are firstly quantized offline. During the inference, any float input will be quantized to an $8$-bit unsigned value before passing to a fixed-point runtime operation, such as QuantizedConv2d, QuantizedAdd, and QuantizedMul, etc. These operations will produce a $32$-bit accumulated result, which will be converted down to an $8$-bit output through an \textit{activation re-quantization} step. Noted that this output will be the input to the next operation.


\subsection{TensorFlow $8$-bit Quantization Scheme}
TensorFlow $8$-bit quantization uses a uniform quantizer, in which all quantization steps are of equal size. Let $x_{float}$ represent for the float value of signal $x$, the TF $8$-bit quantized value, denoted as $x_{quant8}$ can be calculated as:
\begin{equation}\label{eq:quant8}
x_{quant8}=\left[\nicefrac{x_{float}}{\Delta_{x}}\right]-\delta_{x},
\end{equation}

\begin{center}
\begin{equation}\label{eq:delta}
	\Delta_{x}=\frac{x_{max}-x_{min}}{2^b-1} \quad \textrm{and} \quad \delta_{x}=\left[\nicefrac{x_{min}}{\Delta_{x}}\right]
\end{equation}
\end{center}
where $\Delta_{x}$ represents for the quantization step size; $b$ is the bit-width, i.e., $b=8$, and $\delta_{x}$ is the offset value such that float value $0$ is exactly represented. $x_{min}$ and $x_{max}$ are the min and max values of $x$ in the float domain, and $\left[\cdot\right]$ represents for the nearest rounding operation. In the TensorFlow implementation, it is defined as
\begin{equation}\label{eq:rounding}
	\left[x\right] = sgn(x)\cdot\lfloor|x|+0.5\rfloor
\end{equation}
where sgn($x$) is the sign of the signal $x$, and $\lfloor\cdot\rfloor$ represents for the floor operation.

Based on the definitions above, the accumulated result of a convolution operation is computed by:
\begin{equation}\label{eq:accum}
\begin{aligned}
	accum_{float}&= \sum\left( x_{float}\cdot w_{float} \right)\\
	&=\Delta_{x}\Delta_{w} \sum\left(x_{quant8}+\delta_{x}\right)\left(w_{quant8}+\delta_{w}\right)\\
	&=\Delta_{x}\Delta_{w}accum_{int32} \\
\end{aligned}
\end{equation}
Finally, given known min and max values of the output, by combining equation (\ref{eq:quant8}) and (\ref{eq:accum}), the re-quantized output can be calculated by multiplying the accumulated result with $\frac{\Delta_{x}\Delta_{w}}{\Delta_{output}}$, and then subtracting the output offset $\delta_{ouput}$.
\begin{equation}\label{eq:output_quant8}
\begin{aligned}
output_{quant8} &= \left[\frac{1}{\Delta_{output}}accum_{float}\right]-\delta_{ouput} \\
&= \left[\frac{\Delta_{x}\Delta_{w}}{\Delta_{output}}accum_{int32}\right]-\delta_{ouput} \\
\end{aligned}
\end{equation}

\subsection{Metric for Quantization Loss}
As depicted in Figure \ref{fig:quant-pipeline}, there are five types of loss in the fixed-point quantized pipeline, e.g., input quantization loss, weight quantization loss, runtime saturation loss, activation re-quantization loss, and possible clipping loss for certain non-linear operations, such as ReLU6. To better understand the loss contribution that comes from each type, we use Signal-to-Quantization-Noise Ratio (SQNR), defined as the power of the unquantized signal $x$ devided by the power of the quantization error $n$ as a metric to evaluate the quantization accuracy at each layer output.
\begin{equation}\label{eq:sqnr}
SQNR=10\cdot\log_{10}\left(\nicefrac{E(x^2)}{E(n^2)}\right) \quad in \quad dB
\end{equation} 
Since the average magnitude of the input signal $x$ is much larger than the quantization step size $\Delta_{x}$, it is reasonable to assume that the quantization error is zero mean with uniform distribution and the probability density function (PDF) integrates to $1$ \cite{quant-theory}. Therefore, for an $8$-bit linear quantizer, the noise power can be calculated by
\begin{equation}\label{eq:noisepower}
E(n^2)={\int_{-\frac{\Delta_{x}}{2}}^{\frac{\Delta_{x}}{2}}}\frac{1}{\Delta_{x}}n^2dn=\frac{\Delta_x^2}{12}  
\end{equation} 
Substituting equation (\ref{eq:delta}) and (\ref{eq:noisepower}) into equation (\ref{eq:sqnr}), we get
\begin{equation}\label{eq:sqnr2}
SQNR=58.92-10\cdot\log_{10}\frac{(x_{max}-x_{min})^2}{E(x^2)} \quad \text{in} \quad \text{dB}
\end{equation}

SQNR is tightly coupled with signal distribution. From equation (\ref{eq:sqnr2}), it is obvious that SQNR is determined by two terms: the power of the signal $x$, and the quantization range. Therefore, increasing the signal power or decreasing the quantization range can help to increase the output SQNR. 

\subsection{Quantization Loss Analysis on MobileNetV1}
\subsubsection{BatchNorm in Depthwise Convolution Layer}
As shown in Figure \ref{fig:sepconv}(b), a typical MobileNetV1 core layer consists of a depthwise convolution and a pointwise convolution, each of which followed by a Batch Normalization \cite{batchnorm} and a non-linear activation function, respectively. In the TensorFlow implementation, ReLU6  \cite{relu6} is used as the non-linear activation function. Consider a layer input $x=(x^{(1)},...,x^{(d)})$, with $d$-channels and $m$ elements in each channel within a mini-batch, the Batch Normalization Transform in depthwise convolution layer is applied on each channel independently, and can be expressed by,
\begin{equation}\label{eq:batchnorm}
\begin{aligned}
y_{i}^{(k)}&=\gamma^{(k)}\hat{x_{i}}^{(k)} + \beta^{(k)} \\
&=\gamma^{(k)}\frac{x_{i}^{(k)}-\mu^{(k)}}{\sqrt{{\sigma^{(k)}}^2 + \epsilon}} + \beta^{(k)}\\
&\forall i=1,...,m, \quad k=1,...,d
\end{aligned}
\end{equation}
where $\hat{x_{i}}^{(k)}$ represents for the normalized value of $x_{i}^{(k)}$ on channel $k$. $\mu^{(k)}$ and $\sigma^{(k)}$ are mean and variance over the mini-batch. $\gamma^{(k)}$ and $\beta^{(k)}$ are scale and shift. Noted that $\epsilon$ is a given small constant value. In the TensorFlow implementation,  $\epsilon = 0.0010000000475$.

The Batch Normalization Transform can be further folded in the fixed-point pipeline. Let
\begin{equation}\label{eq:batchnorm2}
\alpha^{(k)} = \frac{\gamma^{(k)}}{\sqrt{{\sigma^{(k)}}^2 + \epsilon}} \quad \text{and} \quad \beta'^{(k)} = \beta^{(k)} - \frac{\gamma^{(k)}\mu^{(k)}}{\sqrt{{\sigma^{(k)}}^2 + \epsilon}}
\end{equation}
equation (\ref{eq:batchnorm}) can be reformulated as
\begin{equation}\label{eq:batchnorm3}
\begin{aligned}
&y_{i}^{(k)}=\alpha^{(k)}x_{i}^{(k)} + \beta'^{(k)} \\
&\forall i=1,...,m, \quad k=1,...,d
\end{aligned}
\end{equation}
In the TensorFlow implementation, for each channel $k$, $\alpha$ can be combined with weights and folded into the convolution operations to further reduce the computation cost. 

\begin{figure}[t!]
	\includegraphics[width=8.5cm]{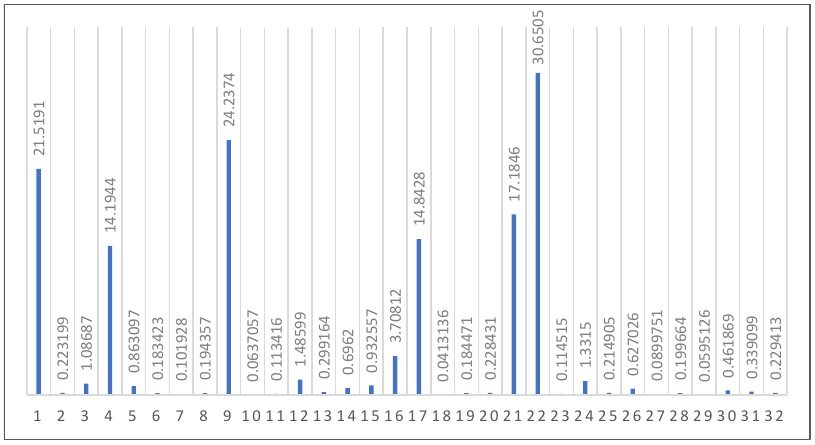}	
	\caption{An example of $\alpha$ values across 32 channels of the first depthwise conv. layer from MobileNetV1 float model}
	\label{fig:alpha-values}
	\centering
\end{figure}

\newcommand{\ssim}{{\sim}}
Depthwise convolution is applied on each channel independently. However, the min and max values used for weights quantization are taken collectively from all channels. An outlier in one channel can easily cause a huge quantization loss for the whole model due to an enlarged data range. Without correlation crossing channels, depthwise convolution may prone to produce all-zero values in one channel, leading to zero variance ($\sigma^{(k)}=0$) for that specific channel. This is commonly observed in MobileNetV1 models. Refer to equation (\ref{eq:batchnorm2}), zero variance of channel $k$ would produce a very large value of $\alpha^{(k)}$ due to the small constant value of $\epsilon$. Figure \ref{fig:alpha-values} shows observed $\alpha$ values across $32$ channels extracted from the first depthwise convolution layer in MobileNetV1 float model. It is noticed that the $6$ outliers of $\alpha$ caused by the zero-variance issue largely increase the quantization range. As a result, the quantization bits are wasted on preserving those large values since they all correspond to all-zero-value channels, while those small $\alpha$ values corresponding to informative channels are not well preserved after quantization, which badly hurts the representation power of the model. From our experiments, without retraining, proper handling the zero-variance issue by changing the variance of a channel with all-zero values to the mean value of variances of the rest of channels in that layer, the top-$1$ accuracy of the quantized MobileNetV1 on ImageNet2012 validation dataset can be dramatically improved from $1.80\%$ to $45.73\%$ on TF8 inference pipeline.

A standard convolution both filters and combines inputs into a new set of outputs in one step. In MobileNetV1, the depthwise separable convolution splits this into two layers, a depthwise separable layer for filtering and a pointwise separable layer for combining \cite{mobilenetv1}, thus drastically reducing computation and model size while preserving feature representations. Based on this principle, we can remove the non-linear operations, \textit{i.e.}, Batch Normalization and ReLU6, between the two layers, and let the network learn proper weights to handle the Batch Normalization Transform directly. This procedure preserves all the feature representations, while making the model quantization-friendly. To further understand per-layer output accuracy of the network, we use SQNR, defined in equation (\ref{eq:sqnr2}) as a metric, to observe the quantization loss in each layer. Figure \ref{fig:sqnr_layeroutput} compares an averaged per-layer output SQNR of the original MobileNetV1 with $\alpha$ folded into convolution weights (black curve) with the one that simply removes Batch Normalization and ReLU6 in all depthwise convolution layers (blue curve).  We still keep the Batch Normalization and ReLU6 in all pointwise convolution layers. $1000$ images are randomly selected from ImageNet2012 validation dataset (one in each class). From our experiment, introducing Batch Normalization and ReLU6 between the depthwise convolution and pointwise convolution largely in fact degrades the per-layer output SQNR.


\subsubsection{ReLU6 or ReLU}
\begin{figure}[t!]
	\centering
	\includegraphics[width=8.5cm]{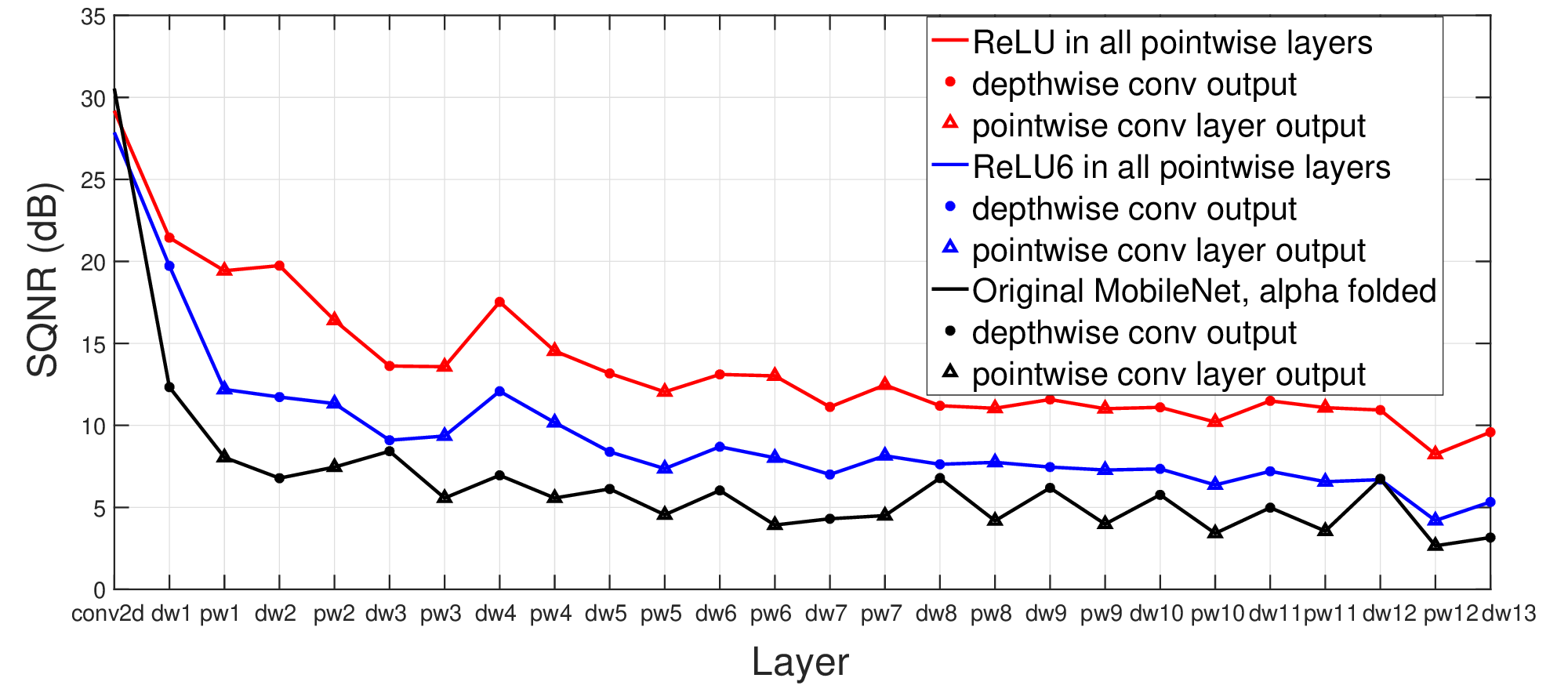}	
	\caption{A comparison on the averaged per-layer output SQNR of MobileNetV1 with different core layer designs}
	\label{fig:sqnr_layeroutput}
\end{figure}

In this section, we still use SQNR as a metric to measure the effect of choosing different activation functions in all pointwise convolution layers. Noted that for a linear quantizer, SQNR is higher when signal distribution is more uniform, and is lower when otherwise. Figure \ref{fig:sqnr_layeroutput} shows an averaged per-layer output SQNR of MobileNetV1 by using ReLU and ReLU6 as different activation functions at all pointwise convolution layers. A huge SQNR drop is observed in the first pointwise convolution layer while using ReLU6. Based on equation (\ref{eq:sqnr2}), although ReLU6 helps to reduce the quantization range, the signal power also gets reduced by the clipping operation. Ideally, this should produce similar SQNR with that of ReLU. However, clipping the signal $x$ at early layers may have a side effect of distorting the signal distribution to make it less quantization friendly, as a result of compensating the clipping loss during training. As we observed, this leads to a large SQNR drop from one layer to the other. Experimental result on the improved accuracy by replacing ReLU6 with ReLU will be shown in Section \ref{sec:experiment}.

\subsubsection{L2 Regularization on Weights}
Since SQNR is tightly coupled with signal distribution, we further enable the L2 regularization on weights in all depthwise convolution layers during the training. The L2 regularization penalizes weights with large magnitudes. Large weights could potentially increase the quantization range, and make the weight distribution less uniform, leading to a large quantization loss. By enforcing a better weights distribution, a quantized model with an increased top-$1$ accuracy can be expected. 

\section{Quantization-Friendly Separable Convolution for MobileNets}
Based on the quantization loss analysis in the previous section, we propose a quantization-friendly separable convolution framework for MobileNets. The goal is to solve the large quantization loss problem so that the quantized model can achieve similar accuracy to the float model while no re-training is required for the fixed-point pipeline.

\subsection{Architecture of the Quantization-friendly Separable Convolution}
Figure \ref{fig:sepconv}(b) shows the separable convolution core layer in the current MobileNetV1 architecture, in which a Batch Normalization and a non-linear activation operation are introduced between the depthwise convolution and the pointwise convolution. From our analysis, due to the nature of depthwise convolution, this architecture would lead to a problematic quantization model. Therefore, in Figure \ref{fig:sepconv}(c), three major changes are made to make the separable convolution core layer quantization-friendly.   

\begin{enumerate}
	\item Batch Normalization and ReLU6 are removed from all depthwise convolution layers. We believe that a separable convolution shall consist of a depthwise convolution followed by a pointwise convolution directly without any non-linear operation between the two. This procedure not only well preserves feature representations, but is also quantization-friendly.  
	\item All ReLU6 are replaced with ReLU in the rest layers. In the TensorFlow implementation of MobileNetV1, ReLU6 is used as the non-linear activation function. However, we think $6$ is a very arbitrary number. Although \cite{relu6} indicates that ReLU6 can encourage a model learn sparse feature earlier, clipping the signal at early layers may lead to a quantization-unfriendly signal distribution, and thus largely decreases the SQNR of the layer output.  
	\item  The L2 Regularization on the weights in all depthwise convolution layers are enabled during the training.
\end{enumerate}

\subsection{A Quantization-Friendly MobileNetV1 Model}
The layer structure of the proposed quantization-friendly MobileNetV1 model is shown in Table\ref{tab:quant-mobilenet}, which follows the overall layer structure defined in \cite{mobilenetv1}. The separable convolution core layer has been replaced with the quantization-friendly version as described in the previous section. This model still inherits the efficiency in terms of the computational cost and model size, while achieves high precision for fixed-point processor.
\begin{table}
	\captionof{table}{Quantization-friendly modified MobileNetV1} \label{tab:quant-mobilenet}
	\renewcommand{\arraystretch}{1.1}
	\begin{tabular}{ |m{1.8cm}|m{3cm}|m{1cm}|m{1cm}| }			
		\hline
		Input 		& Operator 		& Repeat		& Stride 	\\ 	\hline\hline
		224x224x3 	& Conv2d+ReLU 	& 1 			& 2 		\\ 	\hline
		112x112x32 	& DC+PC+BN+ReLU & 1 			& 1  		\\ 	\hline
		112x112x64  & DC+PC+BN+ReLU & 1 			& 2  		\\ 	\hline
		56x56x128 	& DC+PC+BN+ReLU & 1 			& 1  		\\ 	\hline
		56x56x128   & DC+PC+BN+ReLU	& 1 			& 2  		\\ 	\hline
		28x28x256 	& DC+PC+BN+ReLU & 1 			& 1  		\\	\hline
		28x28x256 	& DC+PC+BN+ReLU & 1 			& 2  		\\	\hline
		14x14x512 	& DC+PC+BN+ReLU & 5 			& 1 		\\	\hline
		14x14x512	& DC+PC+BN+ReLU	& 1 			& 2  		\\	\hline
		7x7x1024 	& DC+PC+BN+ReLU & 1 			& 2  		\\	\hline
		7x7x1024 	& AvgPool 		& 1 			& 1 		\\ 	\hline        
		1x1x1024 	& Conv2d+ReLU 	& 1			 	& 1 		\\  \hline
		1x1x1000 	& Softmax 		& 1 			& 1 		\\ 	\hline        
	\end{tabular}
\end{table} 

\section{Experimental Results}\label{sec:experiment}

We train the proposed quantization-friendly MobileNetV1 float models using the TensorFlow training framework. We follow the same training hyperparameters as MobileNetV1 except that we use one Nvidia GeForce GTX TITAN X card and a batch size of 128 is used during the training. ImageNet2012 dataset is used for training and validation. Note that the training is only required for float models.

\begin{figure}[!h]
	\centering
	\includegraphics[width=8.5cm]{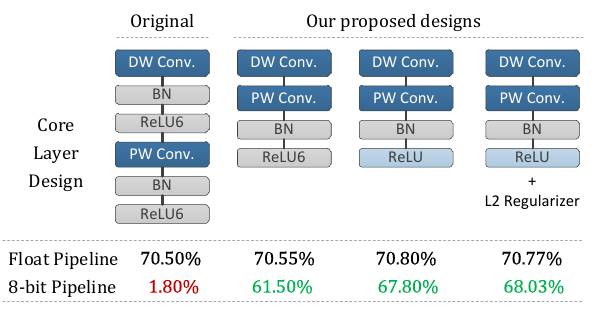}	
	\caption{Top-1 accuracy with different core layer designs on ImageNet2012 validation dataset }
	\label{fig:imagenet_result}
\end{figure}

The experimental results on taking each change into the original MobileNetV1 model in both the float pipeline and the $8$-bit quantized pipeline are shown in Figure \ref{fig:imagenet_result}. In the float pipeline, our trained float model achieves similar top-$1$ accuracy as the original MobileNetV1 TF model. In the $8$-bit pipeline, by removing the Batch Normalization and ReLU6 in all depthwise convolution layers, the top-$1$ accuracy of the quantized model can be dramatically improved from  $1.80\%$ to $61.50\%$. In addition, by simply replacing ReLU6 with ReLU, the top-$1$ accuracy of $8$-bit quantized inference can be further improved to $67.80\%$. Furthermore, by enabling the L2 regularization on weights in all depthwise convolution layers during the training, the overall accuracy of the $8$-bit pipeline can be improved by another $0.23\%$. From our experiments, the proposed quantization-friendly MobileNetV1 model achieves an accuracy of $68.03\%$ in the $8$-bit quantized pipeline, while maintaining an accuracy of $70.77\%$ in the float pipeline for the same model. 

\section{Conclusion and Future Work}
We proposed an effective quantization-friendly separable convolution architecture, and integrated it into  MobileNets for image classification. Without reducing the accuracy in the float pipeline, our proposed architecture shows a significant accuracy boost in the $8$-bit quantized pipeline. To generalize this architecture, we will keep applying it on more networks based on separable convolution, e.g., MobileNetV2 \cite{mobilenetv2}, ShuffleNet \cite{shufflenet} and verify their fixed-point inference accuracy. Also, we will apply proposed architecture to object detection and instance segmentation applications. And we will measure the power and latency with the proposed quantization friendly MobileNets on device.





\begin{thebibliography}{9}
\bibitem{mobilenetv1} 
A. Howard, M. Zhu, B. Chen, D. Kalenichenko, W. Wang, T. Weyand, M. Andreetto, and H. Adam. 
\textit{Mobilenets: Efficient convolutional neural networks for mobile vision applications}.
Apr. 17, 2017, https://arxiv.org/abs/1704.04861.


\bibitem{VGGNet}
K. Simonyan and A. Zisserman. 
\textit{Very deep convolutional networks for large-scale image recognition}. 
Sep.4, 2014, https://arxiv.org/abs/1409.1556.

\bibitem{GoogleNet}
C. Szegedy, W. Liu, Y. Jia, P. Sermanet, S. Reed, D. Anguelov, D. Erhan, V. Vanhoucke, and A. Rabinovich.
\textit{Going deeper with convolutions}. 
In Proceedings of the IEEE Conference on CVPR, pages $1$-$9$, 2015. 1

\bibitem{ResNet}
K. He, X. Zhang, S. Ren, and J. Sun. 
\textit{Deep residual learning for image recognition}.
Dec. 10, 2015, https://arxiv.org/abs/1512.03385.

\bibitem{quanttraining} 
B. Jacob., S Kligys, B. Chen, M. Zhu, M. Tang, A. Howard, H. Adam, and D. Kalengichenko.
\textit{Quantization and Training of Neural Networks for Efficient Integer-Arithmetic-Only Inference}.
Dec.15, 2017, https://arxiv.org/abs/1712.05877.

\bibitem{mobilenetmodel}
\textit{Google TensorFlow MobileNetV1 Model}. https://storage.googleapis.com/download.tensorflow.org/models/tflite/-mobilenet\_v1\_1.0\_224\_float\_2017\_11\_08.zip

\bibitem{gv3}
\textit{Google TensorFlow InceptionV3 Model}. http://download.tensorflow.org/models/-inception\_v3\_2016\_08\_28.tar.gz

\bibitem{tf}
\textit{Google TensorFlow Framework}. https://www.tensorflow.org/

\bibitem{batchnorm}
S. Loff, and C. Szegedy.
\textit{Batch Normalization: Accelerating Deep Network Training by Reducing Internal Covariate Shift}.
Feb. 11, 2015, https://arxiv.org/abs/1502.

\bibitem{quant-theory}
Udo Zölzer.
\textit{Digital Audio Signal Processing , Chapter 2}
John Wiley \& Sons, Dec. 15, 1997

\bibitem{relu6}
A. Krizhevsky.
\textit{Convolutional Deep Belief Networks on CIFAR-10}.
http://www.cs.utoronto.ca/~kriz/conv-cifar10-aug2010.pdf

\bibitem{mobilenetv2}
M. Sandler, A. Howard, M. Zhu, A. Zhmoginov, and L. Chen. 
\textit{Inverted Residuals and Linear Bottlenecks: Mobile Networks for Classification, Detection and Segmentation}.
Jan. 13, 2018, https://arxiv.org/abs/1801.04381.

\bibitem{shufflenet}
X. Zhang, X. Zhou, M. Lin, and J. Sun.
\textit{ShuffleNet: An Extremely Efficient Convolutional Neural Network for Mobile Devices}
Dec. 7, 2017, https://arxiv.org/abs/1707.01083.

\bibitem{dl-compression-survey}
J. Cheng, P. Wang, G. Li, Q. Hu, and H. Lu.
\textit{Recent Advances in Efficient Computation of Deep Convolutional Neural Networks}
Feb. 11, 2018, https://arxiv.org/abs/1802.00939.


\end{thebibliography}
\end{document}